\documentclass[10pt,twocolumn,letterpaper]{article}

\usepackage{cvpr}              

\definecolor{cvprblue}{rgb}{0.21,0.49,0.74}
\usepackage[pagebackref,breaklinks,colorlinks]{hyperref}
\usepackage{fontawesome5} 
\usepackage{hyperref}
\usepackage{minted}
\def\paperID{*****} 
\def\confName{CVPR}
\def\confYear{2026}
\usepackage[table,xcdraw]{xcolor}
\usepackage{booktabs,tabularx,threeparttable,array,makecell}
\usepackage{tabularx}
\usepackage{graphicx}
\usepackage{pifont}
\usepackage{caption}

\usepackage{dblfloatfix} 
\usepackage{siunitx}
\usepackage{threeparttable}
\usepackage{multirow}
\usepackage{float}
\usepackage{adjustbox}
\usepackage{colortbl}
\definecolor{oursrow}{RGB}{235,229,247}
\definecolor{Rule}{HTML}{9AA4B2}
\arrayrulecolor{Rule}

\usepackage[utf8]{inputenc}
\usepackage{amsmath}
\usepackage{amssymb} 
\usepackage[ruled,vlined,linesnumbered]{algorithm2e}
\usepackage{fontawesome5} 

\usepackage{listings}
\usepackage{xcolor}
\usepackage[most]{tcolorbox}
\tcbuselibrary{listings}

\lstdefinestyle{jsonstyle}{
  basicstyle=\ttfamily\small,
  numbers=left,
  numberstyle=\tiny,
  stepnumber=1,
  numbersep=5pt,
  breaklines=true,
  breakindent=1em,
  frame=none,
  showstringspaces=false,
}

\newtcblisting{jsonbox}{
  listing only,
  listing options={style=jsonstyle},
  colback=black!1,
  colframe=black!30,
  boxrule=0.4pt,
  arc=1.5pt,
  left=4pt,right=4pt,top=4pt,bottom=4pt,
  enhanced,
}

\usepackage{makecell}
\usepackage{colortbl}
\definecolor{Header}{HTML}{0F172A}   
\definecolor{Rule}{HTML}{9AA4B2}     

\rowcolors{3}{gray!2}{white}
\arrayrulecolor{Rule}
\newcolumntype{C}[1]{>{\centering\arraybackslash}p{#1}}
\newcolumntype{Y}{>{\centering\arraybackslash}X}

\definecolor{headbg}{RGB}{246,248,252}   
\definecolor{sectbg}{RGB}{244,244,246}   
\definecolor{memcol}{RGB}{232,243,255}   
\definecolor{undcol}{RGB}{235,251,242}   
\definecolor{crosscol}{RGB}{255,243,231} 
\title{ \mymodel: Building Long-Term and Multimodal Memory for Agentic AI }

\author{
\textbf{Chunliang Chen}$^{*}$,
\textbf{Ming Guan}$^{*}$, 
\textbf{Xiao Lin}$^{*}$, 
\textbf{Luxi Lin}$^{*}$, 
\textbf{Jiaxu Li}$^{*}$,
\textbf{Qiyi Wang}$^{*}$,\\
\textbf{Xiangyu Chen},
\textbf{Jixiang Luo},
\textbf{Changzhi Sun}${^{\dagger}}$,
\textbf{Dell Zhang}$^{\#}$,
\textbf{Xuelong Li}$^{\#}$
\\
Institute of Artificial Intelligence (TeleAI), China Telecom\\
\small\url{https://github.com/TeleAI-UAGI/TeleMem}
}

\newcommand{\mymodel}{\textsc{TeleMem}\xspace}

\begin{document}
\maketitle
{\let\thefootnote\relax\footnotetext{\noindent* Equal contribution. $^{\dagger}$ Project lead. $^{\#}$ Corresponding authors.}}

\begin{abstract}
Large language models (LLMs) achieve strong performance on many NLP tasks but remain limited in
long-term interactive settings due to finite context windows and degraded recall over extended
histories.
Retrieval-augmented generation (RAG) alleviates this bottleneck, yet conventional pipelines treat
memories as independent fragments and lack principled mechanisms for consolidation, update, and
causal organization, leading to fragmented context and unstable long-horizon reasoning.
We propose \mymodel, a unified long-term and multimodal memory system that organizes memory as
structured and evolvable semantic trajectories.
\mymodel maintains coherent and hallucination-resistant user profiles by extracting only
dialogue-grounded narrative units, and amortizes write costs through a structured writing pipeline
that batches summaries, retrieves related memories, clusters semantically aligned entries, and
performs LLM-based consolidation before persistent storage.
At the retrieval level, memories are organized into a threaded directed acyclic graph (DAG),
enabling dependency-aware closure-based retrieval that reconstructs coherent causal context for
long-horizon reasoning.
In addition, \mymodel integrates a multimodal memory module with a ReAct-style reasoning
framework to support closed-loop observe--think--act reasoning over complex video content.
Extensive experiments demonstrate that \mymodel outperforms the state-of-the-art Mem0 baseline by
19\% in accuracy on the ZH-4O benchmark, while reducing token usage by 43\% and achieving a
$2.1\times$ speedup.
\end{abstract}

\section{Introduction}
\label{sec:intro}
\section{Introduction}

Large language models (LLMs) have demonstrated remarkable performance across a wide range of
natural language processing tasks~\cite{gpt4o,yang2025qwen3,wang2025logic}.
However, their effectiveness in long-term interactive settings remains fundamentally constrained
by the finite context window of Transformer architectures.
Even though recent long-context models can process hundreds of thousands of tokens~\cite{liu277271533comprehensive},
simply enlarging the context window does not resolve the core challenge.
As interaction histories grow, models struggle to allocate attention to distant yet critical
information, leading to degraded recall of user-specific facts and unstable long-horizon
reasoning~\cite{liu2024lost}.
These limitations hinder the deployment of LLM-based agents in scenarios that require persistent
personalization, continuous learning, and traceable decision making.

Beyond capacity, a more fundamental issue lies in how long-term information is represented and
organized.
Effective memory must preserve not only semantic similarity, but also temporal order, causal
dependency, and state evolution across interactions.
Without such structure, retrieved contexts easily become fragmented, omit critical prerequisites,
and induce inconsistent reasoning.

Retrieval-augmented generation (RAG) has emerged as a practical solution to extend effective memory
beyond the native context window by encoding past interactions into vector embeddings and retrieving
relevant entries via semantic search~\cite{pan2025secom,zhao2024longrag,li2024retrieval}.
While effective at scaling storage capacity, conventional RAG systems treat memories as independent
and unordered fragments.
Such designs lack principled mechanisms for updating, consolidating, or deleting previously written
memories, making it difficult to accommodate evolving user states, resolve contradictions, or
preserve causal consistency over long horizons.

Recent work has therefore explored more structured and adaptive memory architectures that extend
RAG with mechanisms for abstraction, forgetting, and dynamic updates~\cite{zhang2025conversational,xu2025contextagenticfileabstraction}.
MemoryBank~\cite{zhong2024memorybank} introduces a decay mechanism inspired by the Ebbinghaus
forgetting curve to prioritize salient information.
Mem0~\cite{chhikara2025mem0} enables meaning-aware memory operations by allowing an LLM to extract
atomic facts and decide whether to add, update, delete, or ignore them in a streaming manner.
Rsum~\cite{wang2025recursively} applies hierarchical summarization to separate short-term details
from long-term abstractions, while Zep~\cite{rasmussen2025zep} organizes dialogue history as a temporal
knowledge graph to capture causal dependencies.
Although these systems significantly advance long-term memory management, several practical
challenges remain.

\textbf{First}, maintaining a consistent and reliable user profile remains difficult.
Many systems rely on large predefined schemas, while real conversations provide sparse and noisy
signals.
This mismatch often induces hallucinated attributes, incomplete fields, and unnecessary structural
complexity.
\textbf{Second}, write efficiency remains low.
Most pipelines invoke retrieval and LLM-based decision making at every dialogue turn, resulting in
frequent datastore writes, excessive API calls, and degraded throughput and latency.
\textbf{Third}, multimodal reasoning capabilities are still limited.
Existing memory systems are predominantly text-centric and struggle to integrate or reason over
visual, auditory, and temporal information, restricting their applicability in realistic embodied
or multimedia environments.

To address these challenges, we propose \mymodel, a unified long-term and multimodal memory system.
\mymodel organizes memory as structured and evolvable semantic trajectories rather than isolated
fragments, enabling dependency-aware reasoning and consistent context restoration.
It extracts dialogue-supported narrative units to avoid schema-driven hallucinations and maintain
compact, reliable user profiles.
At the storage level, \mymodel employs a structured writing pipeline that batches summaries,
retrieves related memories, clusters semantically aligned entries, and performs LLM-based
consolidation before committing them to persistent storage.
This design amortizes write costs by reducing per-turn LLM invocations, de-duplicates fragmented
information, and substantially improves throughput and token efficiency.
At the retrieval level, memories are organized into a threaded directed acyclic graph (DAG),
enabling closure-based retrieval that reconstructs coherent causal context for long-horizon
reasoning.
In addition, \mymodel incorporates a multimodal memory module that converts raw video streams into
event and object memories, and integrates a ReAct-style reasoning framework~\cite{yao2022react} to
support closed-loop observe--think--act reasoning over complex visual content.

In summary, our main contributions are:
\begin{enumerate}
    \item We introduce \mymodel, a unified long-term and multimodal memory framework that maintains
    coherent and hallucination-resistant user profiles through narrative-driven memory extraction
    and structured organization.
    \item We propose a structured memory graph with threaded dependencies and closure-based
    retrieval, enabling dependency-aware context reconstruction and stable long-horizon reasoning.
    \item We design an efficient memory writing pipeline that performs batched retrieval, semantic
    clustering, and LLM-based consolidation, substantially reducing storage redundancy, token
    overhead, and end-to-end latency.
    \item We develop a multimodal memory module with ReAct-style reasoning, enabling end-to-end
    observe--think--act capabilities for complex video understanding.
    \item Extensive experiments demonstrate that \mymodel outperforms the state-of-the-art Mem0
    baseline by 19\% in accuracy on the ZH-4O benchmark, while reducing token usage by 43\% and
    achieving a $2.1\times$ speedup.
\end{enumerate}

\section{\mymodel System}
\label{sec:system}

\paragraph{Overview.}
Long-horizon agents require memory that preserves not only semantic similarity but also the continuity of experience, including temporal order, causal dependency, and state evolution.
Conventional retrieval-augmented generation (RAG) systems store memories as independent vector chunks, which often leads to fragmented context, missing prerequisites, and unstable reasoning when interactions span long time horizons.
To support stable accumulation and continuous learning, memory must be \emph{threaded} into coherent trajectories rather than retrieved as isolated fragments.

\begin{figure*}[t]
    \centering
    \includegraphics[width=0.9\textwidth]{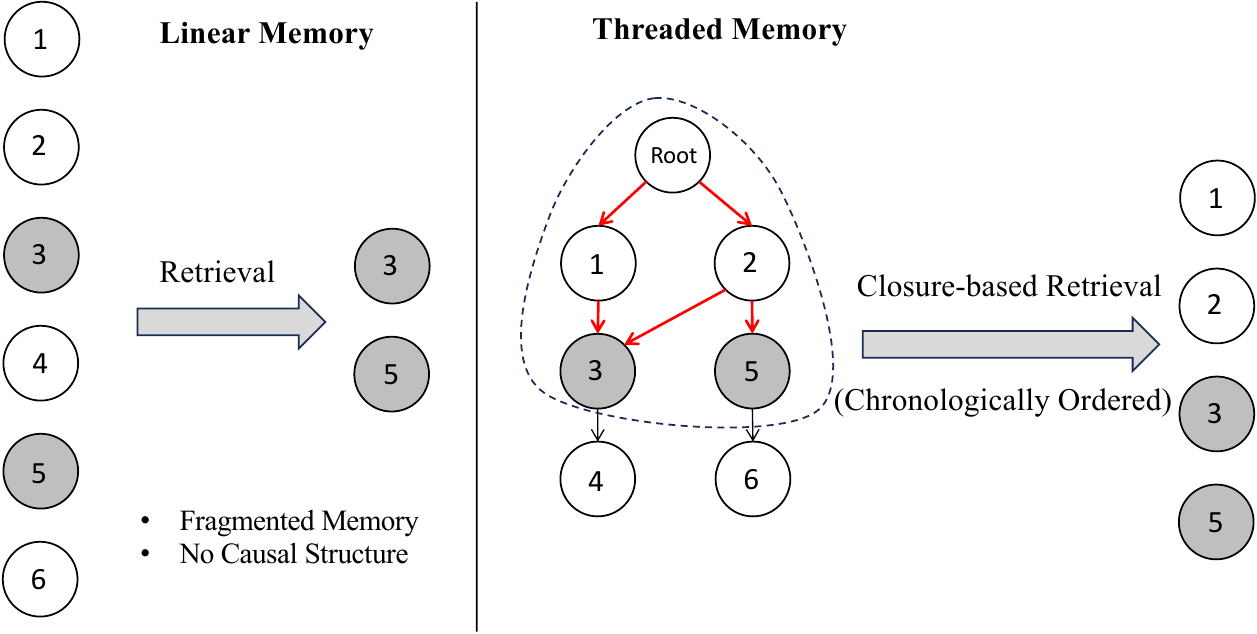}
    \caption{
Comparison between conventional linear memory and the proposed threaded memory with
closure-based retrieval.
\textbf{Left:} Linear memory retrieves isolated entries based on similarity, resulting in
fragmented memories with a lack of causal structure.
\textbf{Right:} Threaded memory organizes memory states as a directed acyclic graph (DAG).
Given a query seed (gray nodes), closure-based retrieval expands along dependency edges to
recover all prerequisite nodes, which are then serialized in chronological order to
reconstruct coherent causal context.
}
    \label{fig:threaded-memory}
\end{figure*}

We address this requirement by organizing long-term memory as a directed acyclic graph (DAG),
where memory states are connected through explicit structural relations to form traceable memory
threads (Sec.~\ref{sec:unified-graph}), upgrading memory from an unordered collection of embeddings
to a structured and evolvable substrate that supports causal reasoning and context restoration.
The overall architecture tightly couples three functional components: (1) a
\emph{representation layer} that converts raw textual and multimodal interactions into semantic
memory states and performs consolidation and updates (Sec.~\ref{sec:dynamic-update}); (2) a
\emph{graph layer} that incrementally organizes these states into a persistent memory graph; and (3) a \emph{memory reading mechanism}
that performs dependency-aware closure-based retrieval to reconstruct coherent causal context for
downstream reasoning (Sec.~\ref{sec:graph-reading}). This unified design supports scalable memory
growth while preserving structural consistency and low-latency access in practice (Fig.~\ref{fig:threaded-memory}).

\subsection{ Memory Graph Formulation}
\label{sec:unified-graph}

We formalize long-term memory as a directed acyclic graph (DAG)
\begin{equation}
G = (V, E),
\end{equation}
where $V$ denotes the set of memory nodes and $E$ denotes directed edges.
Each node $v_i \in V$ is associated with an effective timestamp $\tau(v_i)$,
which reflects the semantic validity order rather than wall-clock time.
All edges satisfy the strict temporal constraint
\begin{equation}
\tau(p) < \tau(v_i),
\end{equation}
which guarantees acyclic evolution.
To avoid isolated memory fragments and ensure global connectivity, we introduce a virtual root node $v_0$ that anchors nodes without explicit predecessors.

\subsubsection{The Node: A Unified Semantic Container}
\label{sec:graph-node}

Each node represents a stabilized semantic memory state produced by the representation layer and serves as a unified container for long-term knowledge.

\paragraph{Semantic categories.}
Nodes may encode heterogeneous semantic states, including:
(1) \emph{profile states}, summarizing stable user and bot attributes (Fig.~\ref{fig:app-text-profile});
(2) \emph{event states}, capturing dynamic interaction-level semantics from textual dialogue or video clips (Fig.~\ref{fig:app-text-event}, Fig.~\ref{fig:app-video-event}); and
(3) \emph{entity and object states}, representing persistent multimodal entities and their temporal evolution (Fig.~\ref{fig:app-video-object}).

\paragraph{Node content.}
Each node stores three components:
(i) a consolidated semantic content representation $\mathrm{Content}(v_i)$,
(ii) an embedding vector $e_i$ for similarity-based retrieval, and
(iii) an effective timestamp $\tau(v_i)$ indicating its semantic validity.
This unified abstraction decouples semantic representation from storage topology, enabling consistent indexing, consolidation, and retrieval across modalities.

\subsubsection{The Edge: Threaded Causal Skeleton}
\label{sec:graph-edge}
Edges define how semantic states are connected and evolve over time.
A directed edge $p \rightarrow v_i$ indicates that node $v_i$ is semantically conditioned on prior node $p$.
Together with the temporal constraint $\tau(p) < \tau(v_i)$, the graph enforces monotonic evolution and prevents circular dependencies.

\paragraph{Minimal causal skeleton.}
To preserve the \emph{threaded} semantics and avoid redundant dependencies,
we maintain the graph as a minimal causal skeleton, where each edge represents an irreducible dependency rather than a shortcut implied by transitive paths.

\paragraph{Pruning criterion.}
An edge $(p \rightarrow v_i)$ is retained if and only if there exists no alternative directed path from $p$ to $v_i$ in the remaining graph:
\[
(p \rightarrow v_i) \in E 
\;\Longleftrightarrow\;
p \not\leadsto v_i \text{ in } G \setminus \{(p \rightarrow v_i)\},
\]
where $a \leadsto b$ denotes reachability via a directed path.
Edges violating this criterion are pruned as transitive or redundant dependencies.
This effectively maintains a transitive reduction of the dependency DAG.
For example, if $pp \rightarrow p$ and $p \rightarrow v_i$ exist, then $pp$ can reach $v_i$ through $p$, and the shortcut edge $pp \rightarrow v_i$ is removed.

\paragraph{Memory threads.}
We define a \emph{memory thread} as any directed path $v_0 \rightsquigarrow v$ in the DAG, where nodes are strictly ordered by increasing effective timestamps.
Such root-to-node paths provide a traceable evolution chain, capturing how topics, states, or beliefs accumulate and change over time as the path progresses chronologically.

By preserving only minimal dependencies under the pruning criterion, the graph avoids transitive shortcuts and keeps these root-to-node threads compact and interpretable.
As a result, threads serve as an explicit causal backbone that supports traceable reasoning, dependency-aware retrieval (e.g., closure construction), and stable long-term knowledge accumulation.

\subsection{Memory Graph Updating}
\label{sec:dynamic-update}

Long-term memory continuously evolves as new interactions arrive and existing memories are refined.
Our system explicitly separates \emph{representation updates} (how semantic node contents are consolidated) from \emph{index updates} (how nodes are organized and connected in the graph), and supports both offline batch processing and online incremental maintenance.

\begin{figure*}[t]
    \centering
    \includegraphics[width=0.9\textwidth]{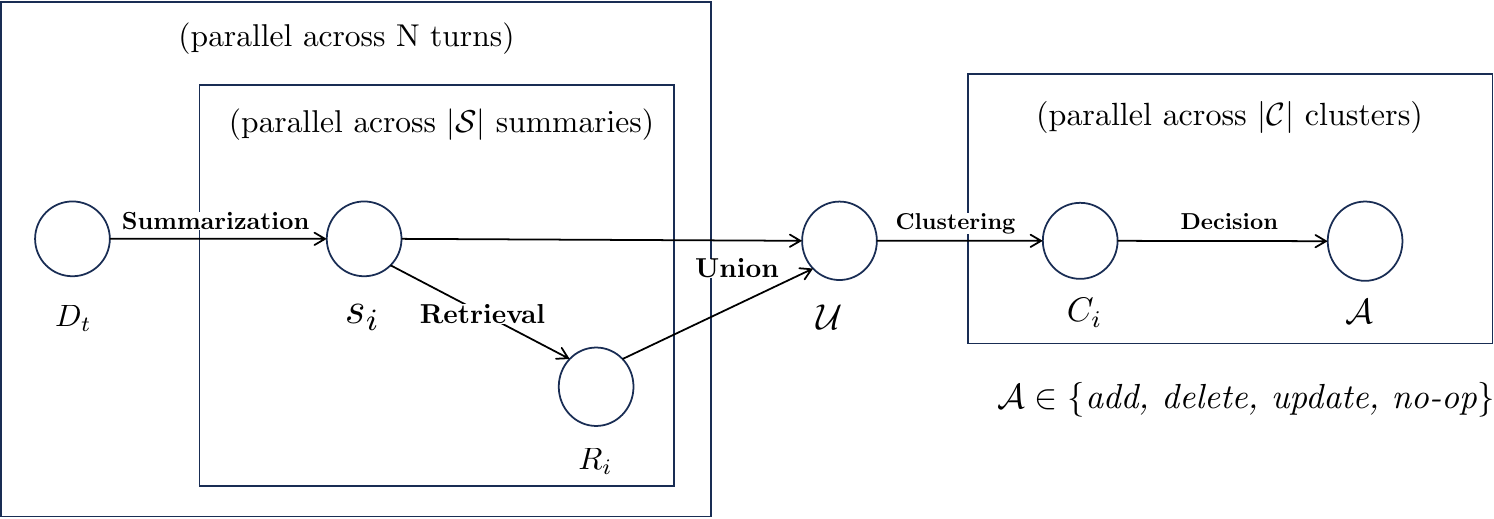}
    \caption{
Overview of the offline node construction pipeline.
A batch of dialogue turns $\{D_t\}_{t=1}^N$ is first summarized in parallel into textual summaries
$s_i$.
Each summary retrieves related historical memories to form retrieval sets $R_i$.
All summaries and retrieved candidates are merged into a unified pool $\mathcal{U}$ and globally
clustered into semantic groups $C_i$.
For each cluster, an LLM performs consolidation and assigns an action
$\mathcal{A} \in \{\emph{add}, \emph{delete}, \emph{update}, \emph{no-op}\}$ to produce stabilized memory updates.
Parallelizable stages across turns, summaries, and clusters are indicated in the diagram.
}
    \label{fig:offline-node}
\end{figure*}

\subsubsection{Graph Update Operators}
\label{sec:graph-ops}

We define two basic operators for maintaining the graph index: \texttt{Insert} and \texttt{ReInsert}.
Both operators enforce the temporal constraint and the pruning criterion (Sec.~\ref{sec:graph-edge}) and share the same edge construction logic: candidate retrieval, pruning, and edge materialization.

\paragraph{\texttt{Insert}}
Given a new node $v_i$ and the current graph $G=(V,E)$, \texttt{Insert}$(v_i,G)$ constructs the incoming edges of $v_i$ through a three-stage procedure: candidate retrieval, redundancy pruning, and edge materialization.

\begin{itemize}[leftmargin=1.2em]

\item \textbf{Candidate retrieval.}
We restrict the parent search space of $v_i$ to historical nodes that satisfy the temporal constraint.
The candidate parent set is defined as the Top-$K$ most similar historical nodes:
\[
\mathcal{H}(v_i) =
\text{Top-}K\!\Big(
\text{sim}\big(v_i,\; \{ v_j \in V \mid \tau(v_j) < \tau(v_i) \} \big)
\Big).
\]
This step bounds parent selection to a local semantic neighborhood and avoids quadratic all-pairs comparison.

\item \textbf{Redundancy pruning.}
To preserve the threaded causal semantics and avoid transitive or dominated dependencies, we retain only \emph{irreducible} parents.
A candidate parent $p \in \mathcal{H}(v_i)$ is kept if and only if there exists no alternative path from $p$ to $v_i$ through another candidate:
\[
\mathcal{P}(v_i)
=
\Big\{
p \in \mathcal{H}(v_i)
\;\Big|\;
\nexists q \in \mathcal{H}(v_i)\setminus\{p\}
\text{ such that }
p \leadsto q
\Big\},
\]
where $p \leadsto q$ denotes reachability in the subgraph induced by
historical nodes $ \{ v_j \in V \mid \tau(v_j) < \tau(v_i) \}$.

\item \textbf{Edge materialization.}
The remaining parents form the incoming edges of $v_i$:
\[
E_{\mathrm{in}}(v_i) = \{ (p \rightarrow v_i) \mid p \in \mathcal{P}(v_i) \}.
\]
These edges are added to the graph index.

\end{itemize}

The \texttt{Insert} operator does not modify any existing node or edge except adding $E_{\mathrm{in}}(v_i)$, and therefore preserves the global DAG property.

\paragraph{\texttt{ReInsert}.}
Given an existing node $v$, \texttt{ReInsert}$(v,G)$ re-attaches $v$ to the graph by recomputing its incoming edges using the same procedure as \texttt{Insert}.
The operator removes all previous incoming edges of $v$ and materializes a new parent set based on its current representation and the pruning criterion.
This local repair is triggered when the semantic content or embedding of $v$ changes, or when its previous dependencies become stale due to upstream updates or deletions.
\texttt{ReInsert} updates only the neighborhood of $v$ and preserves the global DAG property and threaded semantics.

\subsubsection{Offline Batch Updates}
\label{sec:offline-update}

Offline updates are triggered during cold start, large-scale cleanup, or periodic maintenance.
They jointly optimize node representations and graph structure at scale.

\paragraph{Node construction.}
Historical interactions are processed in batches to construct stabilized and compact semantic node representations.
Rather than writing fragmented memories incrementally, the system jointly consolidates multiple dialogue and multimodal segments to reduce redundancy, resolve inconsistencies, and produce high-quality long-term representations.

Let $\mathcal{M}$ denote the current memory store, i.e., the set of existing graph nodes equipped with their embedding index.
Assuming that the system receives a batch of dialogue turns $\{ D_t \}_{t=1}^N$, the representation layer applies a four-stage writing pipeline consisting of \emph{summarization, retrieval alignment, global clustering, and LLM-based consolidation} (Fig.~\ref{fig:offline-node}):

\begin{itemize}[leftmargin=1.2em]
    \item \textbf{Summarization (parallel across turns).}
    Each dialogue turn $D_t$ is independently summarized into one or more textual summaries, denoted as $S_t$.
    The collection of all summaries from the batch is
    \[
        \mathcal{S} = \bigcup_{t = 1}^N S_t .
    \]

    \item \textbf{Retrieval alignment (parallel across summaries).}
    For each summary $s_i \in \mathcal{S}$, the system performs vector retrieval against the memory store $\mathcal{M}$ to identify the top-$k$ most related existing nodes:
    \[
        R_i = \text{Top-}k\!\left( \text{sim}(s_i, \mathcal{M}) \right).
    \]
    This step aligns new content with historical memory to expose semantic redundancy and potential updates.

    \item \textbf{Global semantic clustering (non-parallel).}
    All new summaries and retrieved candidates are merged into a unified candidate pool
    \[
        \mathcal{U} = \left( \bigcup_i R_i \right) \cup \mathcal{S},
    \]
    which is globally clustered into semantic groups
    \[
        \mathcal{C} = \{ C_1, C_2, \ldots, C_m \}, \quad C_j \subseteq \mathcal{U}.
    \]
    Each cluster aggregates semantically related content across both new and historical memory.

    \item \textbf{LLM-based consolidation (parallel across clusters).}
    For each cluster $C_j$, entries are temporally ordered and passed to an LLM, which determines an action
    \[
        \mathcal{A} \in \{ \emph{add}, \emph{delete}, \emph{update}, \emph{no-op} \}
    \]
    for each item.
    This step resolves redundancy and inconsistency and produces refined semantic representations that are re-embedded and emitted as stabilized graph nodes.
\end{itemize}

\paragraph{Edge construction.}
Offline edge construction builds the graph index by applying the \texttt{Insert} operator (Sec.~\ref{sec:graph-ops}) to all nodes in $V$.
Nodes are processed in non-decreasing timestamp order.
For each node $v_i$, \texttt{Insert}$(v_i,G)$ determines its incoming edges based only on historical nodes and does not modify any existing edges.
Therefore, all insertions can be executed independently and in parallel using a read-only vector index.

All locally constructed edges are aggregated to form the final graph $G=(V,E)$.
This parallel construction avoids quadratic all-pairs comparison while preserving temporal validity and the pruning criterion, enabling scalable offline rebuilding.

\subsubsection{Online Incremental Updates}
\label{sec:online-update}

Online updates handle real-time interaction with low latency by incrementally updating both (i) the \emph{node representation} (content/embedding) and (ii) the \emph{graph index} (incoming edges).

\paragraph{Node incremental update.}
In the online setting, node construction follows the same abstraction pipeline as offline consolidation, but operates on a single interaction instance (i.e., batch size $=1$) and omits global clustering for low latency.
Given a new dialogue turn or multimodal observation $D_t$, the representation layer performs:
\begin{itemize}[leftmargin=1.2em]
    \item \textbf{Summarization.}
    $D_t$ is summarized into one or more \emph{candidate semantic nodes}, capturing distinct facts, events, or state changes expressed in the turn.
    
    \item \textbf{Retrieval alignment.}
    Each candidate node is matched against the current memory store to identify highly related existing nodes, exposing potential redundancy or update targets.
    
    \item \textbf{LLM-based decision.}
    Based on the candidate content and retrieved context, an LLM decides whether to \emph{add} a new node, \emph{update} an existing node, or perform \emph{no-op}.
\end{itemize}

This lightweight pipeline enables real-time semantic state extraction while maintaining consistency with the offline consolidation logic.

\paragraph{Edge incremental update.}
Online edge maintenance updates the graph index by invoking the \texttt{Insert} and \texttt{ReInsert} operators (Sec.~\ref{sec:graph-ops}), while preserving the temporal constraint and the pruning criterion (Sec.~\ref{sec:graph-edge}).

\begin{itemize}[leftmargin=1.2em]

\item \textbf{Add.}
A newly created node $v$ is attached to the graph by invoking
\texttt{Insert}$(v, G)$.
By the temporal constraint, $v$ can only depend on historical nodes; therefore, no existing node needs to be modified.

\item \textbf{Update.}
When an existing node $v$ is updated, its semantic representation or embedding may change.
We invoke \texttt{ReInsert}$(v, G)$ to recompute and replace its incoming edges.
To preserve threaded consistency in the downstream structure, all direct children $c \in \mathrm{Ch}(v)$ are additionally refreshed by calling \texttt{ReInsert}$(c, G)$.
Although these children remain temporally valid, their optimal attachment points under the pruning criterion may change after $v$ is revised.

\item \textbf{Delete.}
When a node $v$ is deleted, it is marked as a tombstone and its children $\mathrm{Ch}(v)$ are collected as orphans.
Each orphaned child $c$ is repaired by invoking \texttt{ReInsert}$(c, G)$, which re-attaches $c$ to alternative valid historical parents.
This prevents memory threads from breaking due to the removal of an intermediate dependency.

\end{itemize}

\paragraph{Remark.}
All online updates perform bounded local modifications suitable for low-latency interaction, while periodic offline rebuilding re-optimizes the topology.
This process amortizes local approximation errors and preserves long-term structural consistency of the memory graph.

\subsection{Memory Reading}
\label{sec:graph-reading}

Storing memory well is not enough; it must also be retrieved in a way that restores causal context.
We therefore replace ``Top-$K$ fragment stitching'' with \textbf{closure-based retrieval}, which constructs a minimal closed subgraph that contains both the most relevant evidence and its necessary prerequisites.

\subsubsection{Retrieval Paradigm: Minimal Closed Subgraph}
\label{sec:closure}

Given a query $q$, the system retrieves a closed subgraph that contains the most relevant memory threads together with all their required historical dependencies, while avoiding irrelevant sibling branches whenever possible.

\paragraph{Seed identification.}
We first retrieve a small set of seed nodes by Top-$K$ embedding similarity between $q$ and graph nodes.
If no confident match exists, the virtual root node is used as a fallback seed to preserve global connectivity.

\paragraph{Closure expansion.}
Starting from seed nodes, the system recursively traverses dependency edges backward to collect all reachable ancestor nodes until the virtual root node is reached.
This produces a closed subgraph in which every node’s prerequisites are included.
In practice, if the closure grows excessively large, the traversal can be bounded by a maximum depth or guided by lightweight heuristics (e.g., relevance filtering or budgeted expansion) to control latency and context size.

\paragraph{Context linearization.}
All nodes in the closure are sorted by timestamp and serialized into a linear context sequence, optionally augmented with lightweight structural markers, and injected into the LLM prompt for downstream reasoning.

By reconstructing complete causal threads rather than sampling isolated memory fragments, closure-based retrieval preserves prerequisite consistency, reduces thread confusion, and improves long-horizon reasoning stability.

\subsubsection{Reasoning Paradigm: ReAct-Style Multimodal Agent}
\label{sec:react-on-graph}

For complex multimodal queries, memory reading is performed by a ReAct-style agent that iteratively executes \emph{think--act--observe} cycles over textual memory and raw video content (Alg.~\ref{alg:multimodal-reading}).

\paragraph{Reasoning.}
Based on the query and the accumulated interaction history, the agent reasons about what information is missing, which hypotheses require verification, and whether additional evidence is needed from memory or video.

\paragraph{Action.}
The agent selects among three tools to refine its knowledge:
\begin{itemize}[leftmargin=1.2em]
    \item \texttt{video.retrieval}: retrieve relevant clip timestamps by vector similarity over textual memory;
    \item \texttt{video.rag}: aggregate top-$k$ textual memory entries into the LLM prompt for grounded summarization;
    \item \texttt{video.qa}: perform vision-language question answering over a specified video clip to extract fine-grained visual details.
\end{itemize}

\paragraph{Observation.}
The outputs returned by the selected tools (retrieved memory entries, localized clips, or visual answers) are appended to the agent history and become the observations for the next reasoning step.

Through iterative think--act--observe cycles, the agent progressively refines its understanding of the multimodal context and produces a grounded response.

\SetKwInput{KwInput}{Input}
\SetKwInput{KwOutput}{Output}

\begin{algorithm}[t]
\caption{ReAct-Style Multimodal Memory Reading}
\label{alg:multimodal-reading}

\KwInput{Initial query $q$, language model $\texttt{LLM}(\cdot)$, multimodal memory $\mathcal{M} = \mathcal{M}_e \cup \mathcal{M}_{\mathrm{obj}}$}
\KwOutput{Response}

Initialize history $H \leftarrow [q]$\;

\While{not exceeding \texttt{max\_iterations}}{

    $(\text{action}, \text{args}) \leftarrow \texttt{LLM}(H)$\;

    \If{\text{action} = \texttt{video.retrieval}}{
        $q \leftarrow \text{args}$\;
        $\text{results} \leftarrow \texttt{video.retrieval}(q, \mathcal{M})$\;
    }
    \uElseIf{\text{action} = \texttt{video.rag}}{
        $q \leftarrow \text{args}$\;
        $\text{results} \leftarrow \texttt{video.rag}(q, \mathcal{M})$\;
    }
    \uElseIf{\text{action} = \texttt{video.qa}}{
        $(q, t_{\text{start}}, t_{\text{end}}) \leftarrow \text{args}$\;
        $\text{results} \leftarrow \texttt{video.qa}(q, t_{\text{start}}, t_{\text{end}})$\;
    }
    \uElseIf{\text{action} = \texttt{finish}}{
        \textbf{break}\;
    }

    $H \leftarrow H \;\Vert\; [(\text{action}, \text{args}, \text{results})]$\;
}

\Return $\texttt{LLM}(H)$\;
\end{algorithm}

\section{Experiments}
\label{sec:exp}

\paragraph{Dataset} 
We conduct experiments on two ultra-long dialogue datasets to assess agent memory capabilities. ZH-4O~\cite{chen2025moom} serves as a Chinese role-playing benchmark comprising 28 authentic human-LLM sessions (avg. 600 turns), annotated with 1,068 multiple-choice probing questions to test memory recall.
Complementarily, we utilize LoCoMo~\cite{maharana2024evaluating} as an English dataset, which consists of 10 long-context sessions paired with 1,540 question-answer pairs covering single-hop, multi-hop, open-domain, and temporal reasoning. Notably, we exclude the LoCoMo adversarial subset due to the absence of ground-truth answers.

\paragraph{Baselines} 
We compare our framework against five baselines, categorized into general paradigms and specialized memory systems.
First, we consider two foundational approaches: \textbf{Long context LLM} utilizes the entire conversation history to establish a full-context reference, while \textbf{RAG} retrieves the top-$k$ semantically relevant segments to augment generation.
Second, we include three advanced memory architectures: \textbf{Memobase}~\cite{memobase2025} maintains structured user profiles and event memories; \textbf{A-Mem}~\cite{xu2025mem} utilizes an agentic framework to autonomously link memory notes; and \textbf{Mem0}~\cite{chhikara2025mem0}, a modular memory system designed for scalable deployment with explicit in-context memory operations.

\paragraph{Implementation}
To ensure a fair and consistent evaluation, we re-implemented all baselines using \textbf{Qwen3-8B}~\cite{yang2025qwen3} (configured in ``no-think'' mode) as the backbone LLM, paired with \textbf{Qwen3-8B-embedding}~\cite{yang2025qwen3} for vector representations. Crucially, a unified prompt template is applied for response generation across all settings to strictly control for variance.

\paragraph{Evaluation}
Our evaluation strategies are tailored to each dataset: for ZH-4O~\cite{chen2025moom}, we quantify memory fidelity via QA Accuracy, where the model selects the single correct option for each multiple-choice query against ground-truth labels.
For LoCoMo~\cite{maharana2024evaluating}, following prior work~\cite{chhikara2025mem0}, we employ an LLM-as-a-Judge ($J$) approach using \textbf{GPT-4o}~\cite{gpt4o}, which assesses the factual accuracy, relevance, and completeness of the generated responses.

\subsection{Results on ZH-4O}
\begin{table}[t]
\centering

\begin{tabular}{l c}
\toprule
\textbf{Method} & \textbf{Overall (\%)} \\
\midrule
RAG & 62.45 \\
Mem0 & 70.20 \\
MOOM & 72.60 \\
A-Mem & 73.78 \\
Memobase & 76.78 \\
Long context LLM & 84.92 \\
\midrule
{\mymodel} & \textbf{86.33} \\
\bottomrule
\end{tabular}
\caption{Performance comparison on the ZH-4O benchmark. The metric represents QA Accuracy (\%) across 1,068 probing questions. The best performance is highlighted in \textbf{bold}.}
\label{tab:zh4o_results}
\end{table}

\paragraph{Main Results}
Table~\ref{tab:zh4o_results} presents the performance of different memory paradigms on the ZH-4O benchmark, revealing clear distinctions in their ability to support long-horizon, multi-turn question answering.
We make the following observations.
\begin{itemize}
    \item \textbf{Retrieval-only methods are insufficient.} 
    RAG achieves the lowest accuracy (62.45\%), indicating that flat semantic retrieval without temporal ordering or relational structure fails to support multi-turn, memory-intensive reasoning required by ZH-4O.

    \item \textbf{Explicit memory mechanisms consistently improve performance.} 
    All memory-augmented approaches outperform RAG, demonstrating the necessity of maintaining persistent and updatable memory states for long-horizon dialogue understanding. Among them, Memobase performs best (76.78\%), likely due to its structured user profiling that aligns well with the role-playing characteristics of the benchmark.

    \item \textbf{Long-context modeling alone has inherent limitations.} 
    The Long Context LLM baseline achieves 84.92\% accuracy by leveraging the full dialogue history, but its reliance on raw context makes it susceptible to noise, redundancy, and attention dilution as interactions grow longer.

    \item \textbf{Coordinated read--write memory yields the best results.} 
    Our proposed \mymodel attains the highest accuracy of \textbf{86.33\%}, outperforming both long-context and prior memory-based architectures. This demonstrates that selectively compressing salient information and enabling context-aware memory access is more effective than unstructured context accumulation or loosely coupled memory modules.
\end{itemize}

\begin{table*}[ht]
\centering
\begin{tabular}{l c c c c c c | c}
\toprule
\textbf{Write $\downarrow$ ~ ~  Read $\rightarrow$} & \textbf{0.6B} & \textbf{1.7B} & \textbf{4B} & \textbf{8B} & \textbf{14B} & \textbf{32B} & \textbf{Avg.} \\
\midrule
\textbf{0.6B} & 52.72 & 64.23 & 73.69 & 75.47 & 74.06 & 72.19 & 68.73 \\
\textbf{1.7B} & 61.14 & 71.54 & 79.12 & 78.56 & 80.81 & 77.72 & 74.82 \\
\textbf{4B} & 63.48 & 74.25 & 81.09 & 83.33 & 84.55 & 83.33 & 78.34 \\
\textbf{8B} & 65.54 & 77.15 & 84.36 & 86.33 & 85.77 & 84.93 & 80.68 \\
\textbf{14B} & 67.04 & 78.28 & 84.55 & 86.33 & 87.55 & 85.49 & 81.54 \\
\textbf{32B} & 68.91 & 80.81 & 83.99 & 86.70 & 86.61 & 85.96 & 82.16 \\
\midrule
\textbf{Avg.} & 63.14 & 74.38 & 81.13 & 82.79 & 83.17 & 81.60 & -- \\
\bottomrule
\end{tabular}
\caption{
Memory read–write scaling law. Performance (QA Accuracy, \%) when using different model sizes for the memory write LLM (rows) and the memory read LLM (columns). Results show consistent performance gains as either component scales up, with larger read–write combinations yielding the strongest performance.
}
\label{tab:scaling_law}
\end{table*}

\paragraph{Read-Write Scaling Law}
Table~\ref{tab:scaling_law} reveals a clear memory read–write scaling law by varying the model sizes of the memory write LLM and the memory read LLM. Overall performance improves monotonically as either component scales up, indicating that both memory writing and memory reading contribute substantially to downstream QA accuracy. However, the gains are not symmetric across the two dimensions. Increasing the size of the write LLM leads to consistent improvements across almost all read settings, suggesting that stronger writers produce more informative, compact, and robust memory representations that benefit readers of different capacities.
Scaling the read LLM yields even more pronounced gains, particularly when paired with medium-to-large write models. This indicates that memory retrieval and reasoning capacity plays a critical role in exploiting stored information, especially when the memory content is sufficiently well-structured. Notably, performance saturates when the reader significantly outscales the writer, implying that retrieval capacity alone cannot compensate for low-quality memory writing.
The best results are achieved when both the write and read LLMs are scaled jointly, highlighting the complementary and interdependent nature of memory writing and reading. These findings suggest that effective long-term memory systems should be designed with coordinated read–write capacity rather than over-optimizing either component in isolation.


\subsection{Results on LoCoMo}
\paragraph{Main Results}
Table~\ref{tab:locomo_results} compares different methods on the LoCoMo benchmark across four reasoning categories. Overall, the Long Context LLM achieves the strongest performance (70.71\%), demonstrating the effectiveness of processing full interaction histories for complex memory-centric tasks. This advantage is particularly pronounced on temporal questions, where access to complete chronological context is crucial. 
In contrast, retrieval-based baselines perform poorly across all categories, highlighting the limitations of static semantic retrieval for long-term conversational reasoning.
Memory systems exhibit diverse strengths across reasoning types. Mem0 performs well on single-hop questions (54.96\%) and temporal reasoning (60.28\%), suggesting that its modular memory abstraction is effective for fact recall and time-sensitive information. 
Memobase excels on multi-hop reasoning (66.04\%), likely benefiting from its structured memory representations that better support compositional inference. A-Mem shows more balanced but moderate performance, indicating that agentic memory control alone is insufficient without strong memory structuring.
\mymodel achieves competitive performance on single-hop (64.53\%) and temporal reasoning (78.47\%), approaching the Long Context LLM in these categories, which suggests that its memory abstraction effectively captures salient factual and temporal information.
However, its weaker performance on multi-hop questions indicates remaining challenges in supporting complex relational reasoning over stored memories.
Overall, these results highlight that different memory mechanisms favor different reasoning patterns, and no single approach uniformly dominates all categories, underscoring the importance of task-aware memory design for long-horizon QA.

\begin{table*}[h]
\centering
\begin{tabular}{l c c c c c}
\toprule
\textbf{Method} & \textbf{Single-Hop} & \textbf{Multi-Hop} & \textbf{Open Domain} & \textbf{Temporal} & \textbf{Overall} \\
\midrule
RAG & 20.91 & 32.39 & 35.41 & 48.03 & 39.03 \\
Mem0 & 54.96 & 31.15 & 40.62 & 60.28 & 52.01 \\
A-Mem & 44.32 & 35.82 & 33.33 & 56.59 & 48.57 \\
Memobase & 50.70 & 66.04 & 37.50 & 58.97 & 57.59 \\
Long context LLM & 64.89 & 47.66 & 42.70 & 84.66 & 70.71 \\
\midrule
\mymodel & 64.53 & 20.56 & 40.62 & 78.47 & 61.49 \\
\bottomrule
\end{tabular}
\caption{Performance comparison on the LoCoMo benchmark. All metrics represent accuracy (\%) evaluated via LLM-as-a-Judge across 1,540 question-answer pairs.}
\label{tab:locomo_results}
\end{table*}

\section{Related Work}
\label{sec:related}

\paragraph{Text LTM} 
Recent research on equipping Large Language Models (LLMs) with text long-term memory has primarily evolved through system-level abstractions, cognitive-inspired mechanisms, and structured agentic frameworks. To transcend fixed context windows, operating system paradigms like MemGPT~\cite{packer2023memgpt} and MemOS~\cite{li2025memos} leverage virtual memory abstractions and hierarchical scheduling to orchestrate data flow between active context and external storage. Complementing these architectural innovations, bio-inspired models such as MemoryBank~\cite{zhong2024memorybank}, LightMem~\cite{fang2025lightmem}, and Nemori~\cite{nan2025nemori} integrate human cognitive theories—ranging from the Ebbinghaus forgetting curve to the Atkinson-Shiffrin model—to dynamically prioritize, segment, and decay information for efficient retrieval. On the structural front, Mem0~\cite{chhikara2025mem0}, A-Mem~\cite{xu2025mem}, and MIRIX~\cite{wang2025mirix} propose  modular systems that enable self-evolving knowledge consolidation and multi-agent coordination.

\paragraph{Graph LTM}
Recent literature on long-term memory for LLM agents highlights a transition from unstructured storage to graph-based architectures that enhance reasoning, consistency, and scalability. Foundational works like ${Mem0}^{\tiny g}$ \cite{chhikara2025mem0} and Zep \cite{rasmussen2025zep} establish production-ready and temporal knowledge graph systems to manage agent memory efficiently. To address the complexity of long-term interactions, G-Memory \cite{G-Memory}, LiCoMemory \cite{huang2025licomemory}, and SGMEM \cite{wu2025sgmem} proposes hierarchical graph structures that organize information at varying granularities, from sentence-level details to high-level insights, thereby facilitating self-evolution and reducing fragmentation.
Enhancing the reasoning capabilities over these structures, GraphCogent \cite{wang2025graphcogent} and D-SMART \cite{D-SMART} integrate working memory models and dynamic reasoning trees to support complex graph understanding and dialogue consistency, MemQ \cite{memq} focuses on optimizing knowledge graph reasoning. Furthermore, \cite{xia2025experience} introduces trainable graph memories that abstract agent trajectories into strategic meta-cognition. Collectively, these studies demonstrate that structured, hierarchical, and dynamic graph memories are essential for developing autonomous agents capable of coherent, long-term strategic planning and reasoning.

\paragraph{Parametric LTM}
Parametric memory aims to encode knowledge or contextual information directly into the weights or persistent hidden states of neural networks~\cite{liu2024generation,liu2025protein,miao2025taso}, distinguishing itself from non-parametric approaches that rely on external vector databases. 
Wang et al.~\cite{wang2023augmenting} propose a decoupled memory mechanism, employing a residual side network to cache long-term context while keeping the base LLM frozen. Similarly, MLPMemory~\cite{wei2025mlp} internalizes retrieval by training multilayer perceptrons to approximate k-nearest neighbor distributions as differentiable mappings. Pushing the internalization of retrieval further, Tay et al.~\cite{tay2022transformer} introduce the Differentiable Search Index, which eliminates external indices by training the model to map queries directly to document identifiers via its parameters. regarding personalized adaptation, Zhang et al.~\cite{zhang2024personalized} combine LoRA with Bayesian optimization to inject dialogue history into model weights. Complementary to these injection methods, Meng et al.~\cite{meng2022locating} explore the interpretability of parametric storage, proposing ROME to locate and directly edit factual associations within the Transformer's MLP layers. Overall, parametric memory methods offer scalability and unified reasoning but face challenges regarding update costs and potential misalignment with the base model.

\paragraph{Multimodal LTM}
Multimodal memory~\cite{wang2024videoagent,zhang2025deep} has drawn increasing attention as modern agents must store and reason over long-horizon visual and textual information. Mem0~\cite{chhikara2025mem0} introduces a multimodal image interface but ultimately reduces visual inputs to captions and continues to operate purely in the textual space. M3-Agent~\cite{long2025seeing} extends memory into the multimodal domain by maintaining entity-level representations across audio–visual streams and enabling agentic retrieval over long video sequences. Inspired by hippocampal mechanisms, HippoMM~\cite{lin2025hippomm} proposes cross-modal event encoding and temporal consolidation to support richer multimodal recall. MemVerse~\cite{liu2025memverse} further explores lifelong multimodal memory through a hierarchical retrieval framework with periodic distillation for compactness. In contrast, VisMem~\cite{yu2025vismem} emphasizes preserving visual latent representations rather than relying solely on textual abstractions. Overall, existing multimodal memory systems highlight the need for multi-granular representations and adaptive retrieval, while our approach advances this direction with a lightweight hierarchical design tailored for continuous multimodal streams.

\section{Conclusion}
\label{sec:conclusion}
We introduced \mymodel, a unified long-term and multimodal memory system that overcomes key limitations of existing RAG-based approaches.
By extracting narrative-grounded information and employing a structured writing pipeline for batching, clustering, and consolidation, \mymodel maintains coherent user profiles while greatly improving storage and token efficiency. 
Its multimodal memory module with ReAct-style reasoning further enables accurate observe–think–act processing for complex video content.
Experiments on the ZH-4O benchmark show that \mymodel substantially outperforms Mem0 in accuracy, efficiency, and speed, underscoring its effectiveness.

\paragraph{Acknowledgments.} 
The development of \mymodel has been greatly influenced by the contributions of open-source communities and prior research efforts.
We express our sincere appreciation to the teams and projects whose work has inspired and supported this system, including Mem0, Memobase, MOOM, DVD, and Memento. 
Their innovations have provided valuable foundations upon which this work builds.

{
    \small
    \bibliographystyle{ieeenat_fullname}
    \bibliography{main}
}

\clearpage
\newpage
\appendix

\section{Example Memory Entries}
\label{app:memory-examples}

This section provides representative JSON snippets illustrating how different types of
memory nodes are serialized in our system.
These examples are intended to clarify the unified schema, field semantics, and provenance
information used by the memory graph, rather than to enumerate all possible attributes.

\begin{figure}[h]
    \centering
    \includegraphics[width=0.85\linewidth,
        trim=40 30 40 60,clip]{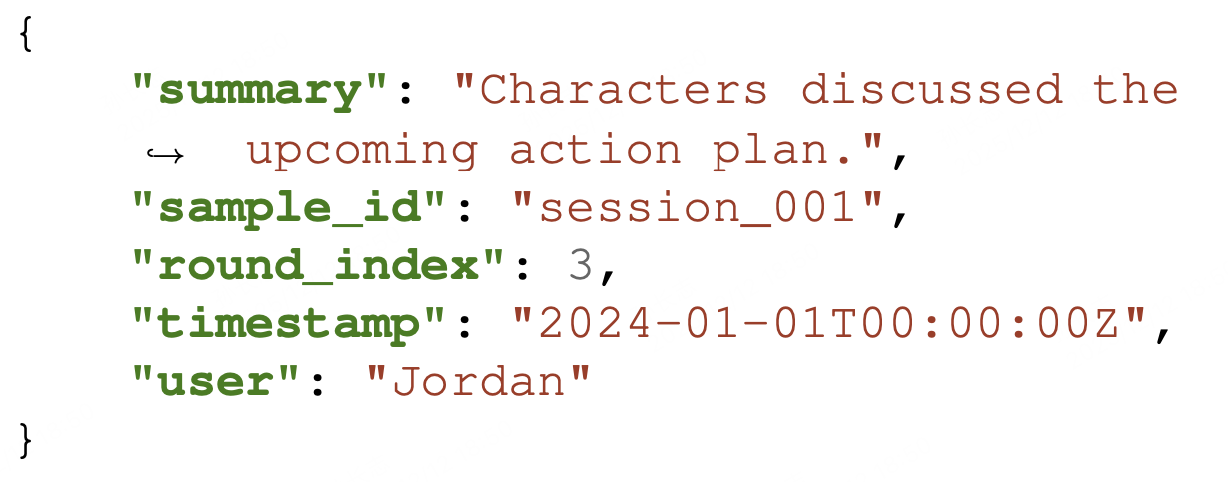}
    \caption{Example JSON snippet of a text profile memory entry.}
    \label{fig:app-text-profile}
\end{figure}
\paragraph{Text profile memory.}
Figure~\ref{fig:app-text-profile} shows an example of a text profile memory entry, which
captures relatively stable attributes of a user or an agent.
Such entries typically include a high-level semantic summary, a persistent identifier,
and lightweight metadata (e.g., timestamps or session indices) that enable long-term
tracking and consolidation across interactions.

\begin{figure}[h]
    \centering
    \includegraphics[width=0.85\linewidth,
        trim=40 30 40 60,clip]{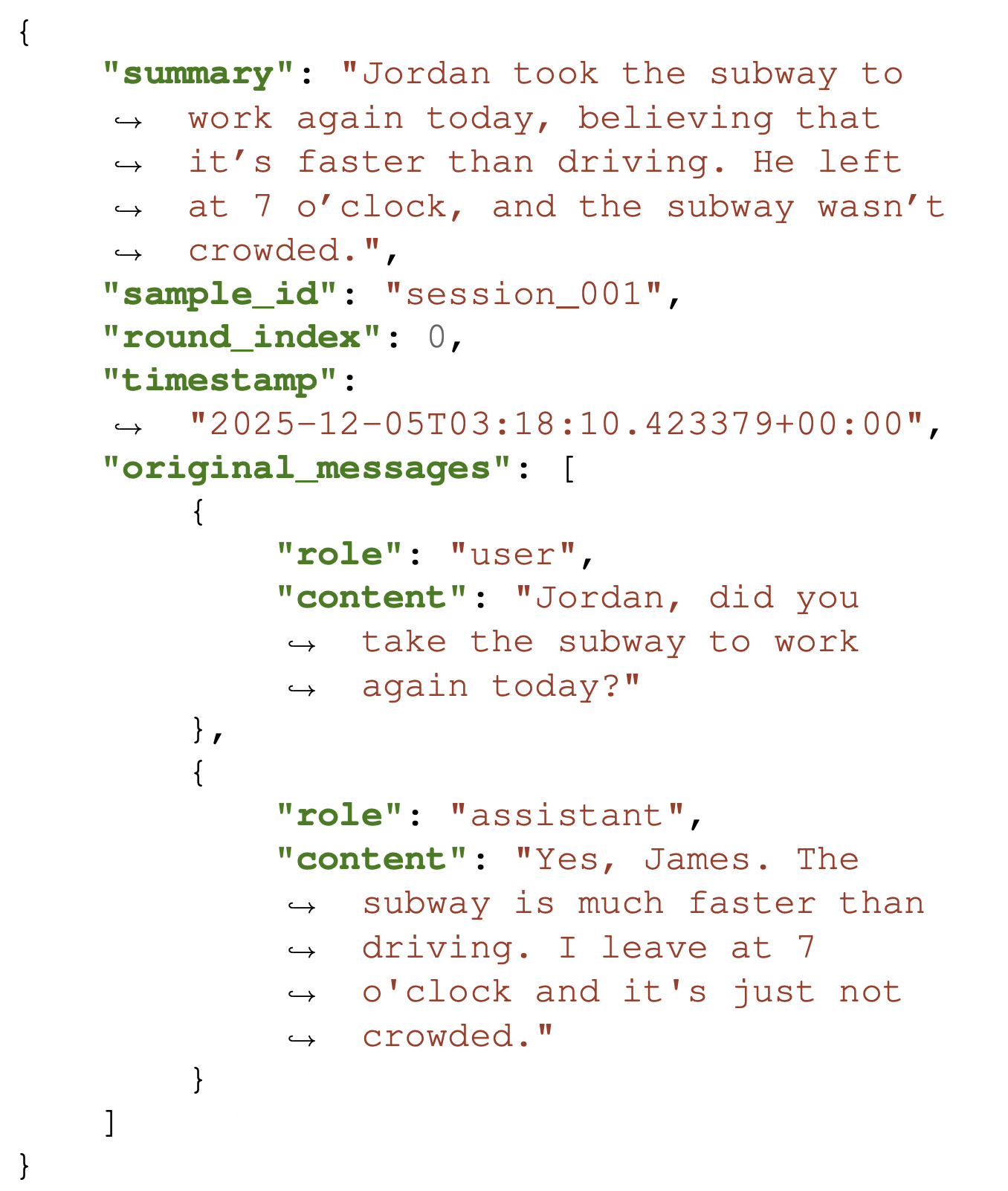}
    \caption{Example JSON snippet of a text event memory entry.}
    \label{fig:app-text-event}
\end{figure}
\paragraph{Text event memory.}
Figure~\ref{fig:app-text-event} illustrates a text event memory entry corresponding to a
single interaction or dialogue turn.
In addition to the summarized semantic content, the entry records the original messages,
roles, and temporal information, enabling traceability back to raw conversational context
and supporting dependency-aware retrieval.

\begin{figure}[h]
    \centering
    \includegraphics[width=0.85\linewidth,
        trim=40 30 40 60,clip]{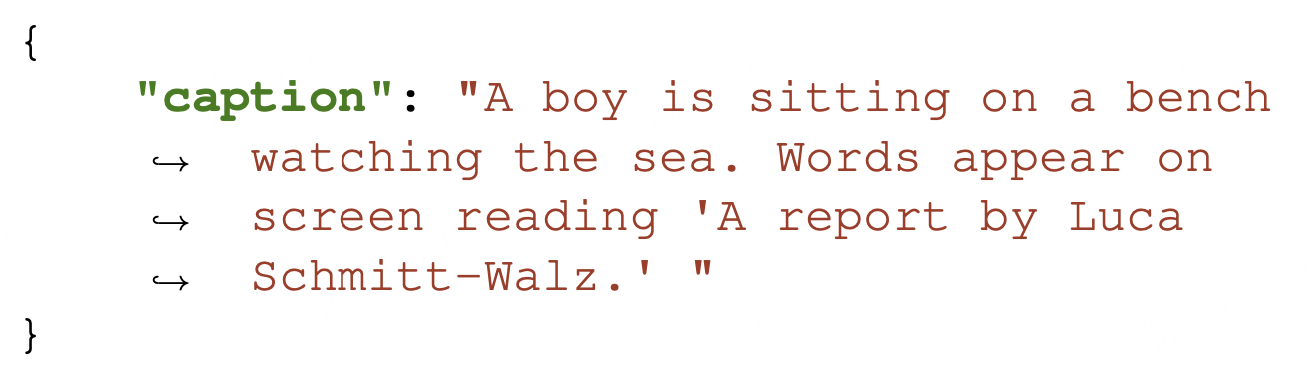}
    \caption{Example JSON snippet of a video event memory entry.}
    \label{fig:app-video-event}
\end{figure}

\paragraph{Video event memory.}
Figure~\ref{fig:app-video-event} presents a video event memory entry that encodes localized
multimodal observations.
Besides the textual caption, such entries may include aligned timestamps, clip references,
or auxiliary annotations, which allow the system to ground high-level semantics in concrete
visual evidence.

\begin{figure}[h]
    \centering
    \includegraphics[width=0.85\linewidth,
        trim=40 30 40 60,clip]{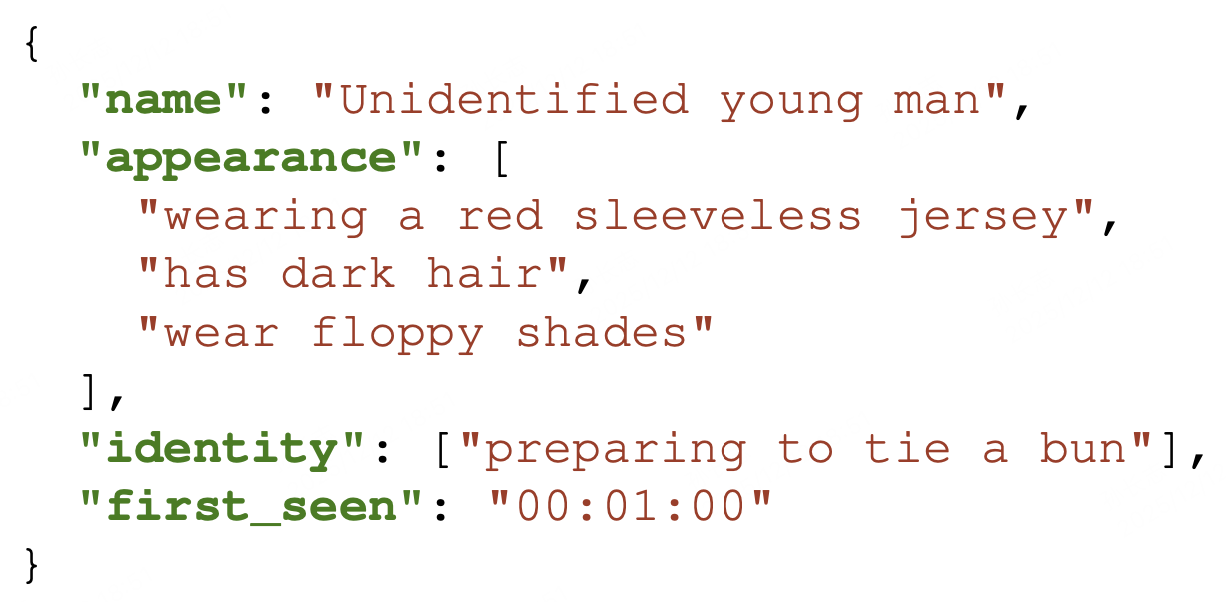}
    \caption{Example JSON snippet of an object memory entry.}
    \label{fig:app-video-object}
\end{figure}

\paragraph{Object memory.}
Figure~\ref{fig:app-video-object} shows an object memory entry representing a persistent
entity tracked across video frames or scenes.
Typical fields include object identity, appearance attributes, behavioral cues, and the
first observed timestamp, supporting long-term entity consistency and temporal reasoning.


\end{document}